\documentclass[10pt,conference]{IEEEtran}
\ifCLASSINFOpdf
  \usepackage[pdftex]{graphicx}
  \graphicspath{{./figures/}}
\else
\fi
\usepackage{amsmath}
\usepackage{caption}
\usepackage{hyperref}

\begin{document}

%
\title{Machine Learning Model Drift Detection Via Weak Data Slices}

\author{\IEEEauthorblockN{Samuel Ackerman\IEEEauthorrefmark{3},
Parijat Dube\IEEEauthorrefmark{2},
Eitan Farchi\IEEEauthorrefmark{1}, 
Orna Raz\IEEEauthorrefmark{1} and
Marcel Zalmanovici\IEEEauthorrefmark{1}}
\IEEEauthorblockA{\IEEEauthorrefmark{3}IBM Research, Haifa, Israel\\
Email: samuel.ackerman@ibm.com}
\IEEEauthorblockA{\IEEEauthorrefmark{1}IBM Research, Haifa, Israel\\
Email: farchi, ornar, marcel @il.ibm.com}
\IEEEauthorblockA{\IEEEauthorrefmark{2}IBM Research, Yorktown Heights, USA\\
Email: pdube@us.ibm.com}
}

\maketitle

\begin{abstract}
Detecting drift in performance of Machine Learning (ML) models is an acknowledged challenge. For ML models to become an integral part of business applications it is essential to detect when an ML model drifts away from acceptable operation. However, it is often the case that actual labels are difficult and expensive to get, for example, because they require expert judgment. Therefore, there is a need for methods that detect likely degradation in ML operation without labels. We propose a method that utilizes feature space rules, called data slices, for drift detection. We provide experimental indications that our method is likely to identify that the ML model will likely change in performance, based on changes in the underlying data.
\end{abstract}


%
\IEEEpeerreviewmaketitle

\section{Introduction} \label{sec:introduction}
AI-Infused Applications (AIIA) are becoming prevalent, yet the statistical nature of their operation makes it challenging to  rely on such applications for business purposes \cite{AIIA}. One major challenge is in being able to infer that a machine learning (ML) model's measured performance, such as its accuracy, is likely to degrade relative to a prior baseline and become potentially unacceptable. 
Even when model performance in the field is initially similar to during training, it might degrade over time, or degrade when the model is re-deployed elsewhere, for example, in a different geography. 


This \emph{ML model performance drift} is often caused by \emph{data drift}, that is, gradual changes in the underlying data.
The challenge is to detect data drift, even when the ground truth or labels are unavailable. In this paper we target the detection of model performance drift and data drift that is likely to cause it, and so we use the term \emph{drift} to indicate both. Detection of drift could proactively trigger various remediation actions. For example, asking a human expert to examine a sample of the field data and compare their labels with those of the ML model, or triggering a timely retraining of the model. 

We propose a method to detect such drift in a test set based on changes in the input or feature space only, ignoring the true or predicted target feature values. We demonstrate its effectiveness in detecting  data changes that are likely to cause data or model performance drift over three data sets.

We utilize previous work that finds \emph{weak data slices} \cite{10.1007/978-3-030-62144-5_6,10.1145/3338906.3340442}. Section \ref{subsec:slicesForDrift} defines weak data slices. In a nutshell, these are data regions where the ML model has a statistically significant higher error rate compared to the average error rate. Moreover, a user can further refine the set of data slices such that they capture important business requirements. For example, the ML model should have an acceptable error rate for input records coming from young people ages 5 to 18 that live in the mid-west or in the east coast.

The main contributions of our work are 
\begin{enumerate}
    \item Providing a feature space method for drift detection including the definition of a data slices-based distribution for drift detection.
    \item Providing initial evidence for the effectiveness of our drift detection method and characterizing the types of data drift it is effective in detecting.
\end{enumerate}

Section~\ref{sec:methodology} details our drift detection method, including more details about its usage of weak data slices, hypothesis testing, and drift detection goals. Section~\ref{sec:experiments} details the experimental settings for validating the effectiveness of our drift detection method. Section~\ref{sec:results} provides the main results. These indicate that both goals of our method are achieved. Finally, Section~\ref{sec:related} summarizes the main related work and Section~\ref{sec:discussion} concludes and outlines directions for future work.

\section{Methodology} \label{sec:methodology}

Our drift detection method utilizes weak data slices as Section \ref{subsec:slicesForDrift} explains. Section \ref{subsec:hypothesisSetup} defines the hypothesis test setup that enables our method to compare two data sets for drift detection. Our drift detection technique targets two goals.  Goal 1 (Section~\ref{subsec:goal1}) is to detect data drift that are likely to affect model performance, as Section  defines. Goal 2 (Section~\ref{subsec:goal2}) is to is to infer likely model performance drift directly, without using label values.

\subsection{Utilizing weak data slices for drift detection} \label{subsec:slicesForDrift}

Assume an ML model $\mathcal{M}$ is trained and returns predictions $\hat{\mathbf{y}}$ on a test dataset $D$ with true target feature values $\mathbf{y}$.  A data slice $S$ on the dataset $D$ is defined in terms of the feature space of $D$ (ignoring the target $\mathbf{y}$).  It is a rule indicating value ranges for numeric features, sets of discrete values for categorical features, and combinations of the above.  The hypothetical slice given in  Section~\ref{sec:introduction} can be denoted $S_1=(5\leq\textrm{age}\leq 18)\: \&\:(\textrm{region}\in\{\textrm{mid-west},\textrm{west coast}\})$, an intersection of subsets of the two features `age' and `region'.  Our drift detection can work with data slices from many sources (e.g., human-provided rather than found algorithmically), as long as they are defined as rules over the feature space, as above. 

A \emph{weak data slice} is a data slice for which the mis-classification rate (MCR) on observations that fall in it is significantly higher than the overall MCR of $\mathcal{M}$ on the given $D$.  Currently, our procedure is only used for classification problems, but the notion of a weak slice can be extended to numeric target prediction, where $\mathcal{M}$ has, say, a higher root mean squared error on the slice than on $D$ overall.

\begin{figure}
    \centering
    \includegraphics[scale=0.25]{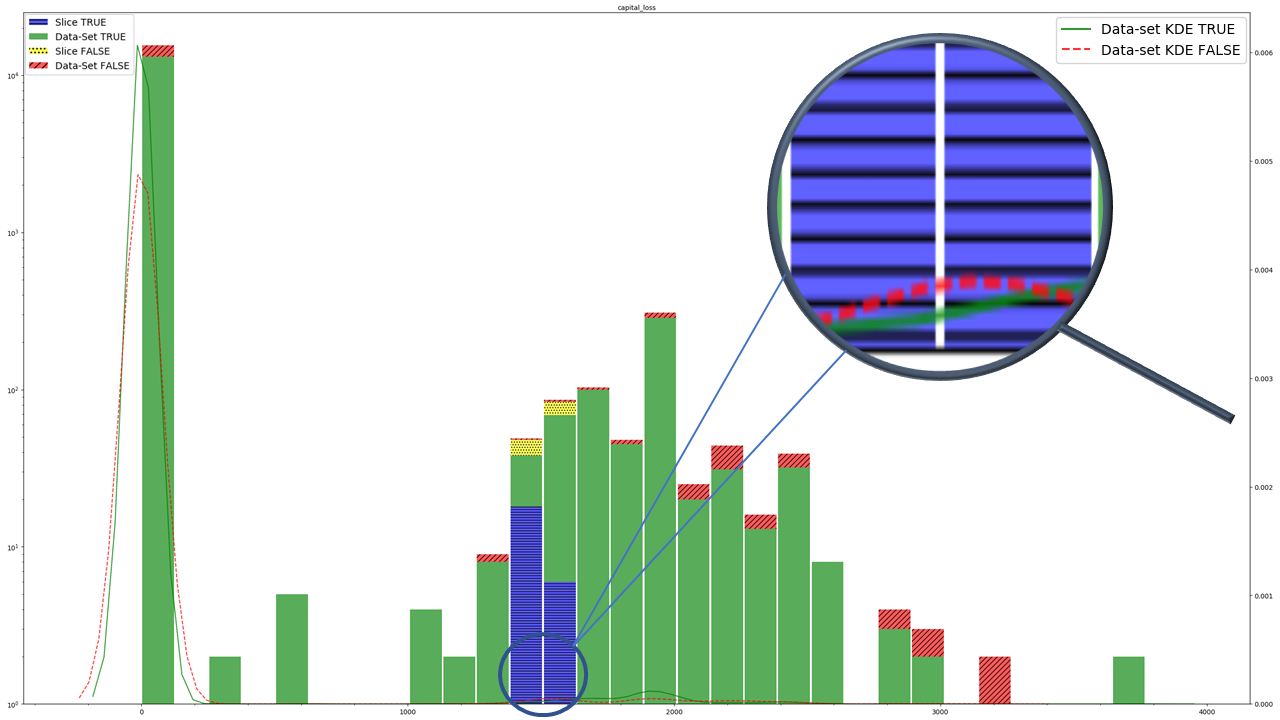}   \caption{\label{fig:sliceExample} An example weak data slice from the Adult data set \ref{subsec:datasets} over the capital loss continuous valued feature. It shows the histogram of the feature with the slice marked on top. The KDE for all the data records (full line) and for the slice (dashed line) are also shown. The y-axis is logarithmic}
\end{figure}

Figure~\ref{fig:sliceExample} shows the empirical distribution (logarithmic axis), represented by the bar heights, of the `capital\_loss' feature from the Adult dataset (see Section~\ref{subsec:datasets}).  The red portions are the fraction of observations in each bin that $
\mathcal{M}$ misclassifies. The solid green and dashed red lines show the kernel density estimates (KDEs) of `capital\_loss' for all and only mis-classified records, respectively.  In the two bins indicated by the circle inset, the blue (correct class) and yellow (incorrect) portions represent observations falling in some slice (range of values within these bins).  The MCR (\{yellow\}/\{yellow + blue\} area proportion) is higher in this slice than elsewhere (\{red\}/\{red + green\}), and hence is a weak slice.  This is shown by the concentration of mistakes overall (red dashed line) being higher than the average observation concentration (sold green line) in the inset.  Although the MCR is also higher in the right tail of the distribution, we do not search for slices there because the data distribution is too sparse to be significant.


Existing work~\cite{10.1007/978-3-030-62144-5_6,10.1145/3338906.3340442} finds weak data slices given a trained ML model $\mathcal{M}$ and a test dataset $D$. The purpose of the data slices work is multi-fold: assist in fault localization, assist in better understanding of the model $\mathcal{M}$'s behavior and in determining whether it is acceptable or not, as well as direct remediation actions.  
Although a data slice in general can be a conjunction of any number of features of $D$, in practice, we typically only consider up to 3-way feature combinations.  Higher-order combinations are both more computationally-intensive to find and may be of less practical use.  Our observation in this work is that weak data slices can be used to define an empirical distribution for drift detection, beyond simply assisting in model diagnosis. Hereafter in this work, `slices'  refer specifically to \emph{weak ones}.

Weak data slices found will differ in their importance and utility in diagnosing
$\mathcal{M}$'s performance.  All other things equal, slices are more useful the higher their average error rate, the number of records, the statistical significance of the slice, and potentially the uniqueness of the problematic observations in the slice. In ongoing research, not discussed here, we consider a heuristic to use these attributes to give slices a rank score, quantifying their potential utility to a user.



\subsection{Problem formulation and motivation}
\label{subsec:guidingMotivation}

Slices represent regions of the feature space of $D$ over which $\mathcal{M}$ tends to err, and thus, intuitively, the larger the slices on a given dataset, the more errors $\mathcal{M}$ is likely to have.  Mapping a slice rule $S_i$ to a specific dataset $D$ (possibly different from the one used to define $S_i$) means determining the subset of observations in $D$ that satisfy the feature constraint $S_i$.  When $S_i$ is mapped to a dataset $D$, two attributes of the mapped observation subset are of particular interest: 
\begin{itemize} 
    \item size $n_i$: the number of records in the mapped subset.
    \item $m_i$: the number of mis-classified records falling under the rule.  The slice's MCR is thus $m_i/n_i$.
\end{itemize}

In our applications, we use the slices to nonparametrically compare two datasets $D_1$ and $D_2$ of the same feature set; dataset $j,\: j=1,2$ has $N_j$ observations, of which $M_j$ are mis-classified.  First, the slice-finder algorithm is run on $D_1$, on which it is known whether each observation is classified correctly.  This yields $K$ slice rules $S=\{S_1,\dots,S_K\}$, for which slice $i,\:i=1,\dots,K$ has size $n_{1,i}$ and $m_{1,i}$ mis-classifications. The MCR on each slice is $m_{1,i}/n_{1,i}$. Because the method finds slices with above-average MCRs, we have $m_{1,i}/n_{1,i} > M_1/N_1$ for each $i$. 

Mapping rules $\{S_i\}$ to dataset $D_2$'s feature values yields $K$ slices with corresponding sizes $n_{2,i},\: i=1,\dots, K$.  The work in~\cite{10.1007/978-3-030-62144-5_6,10.1145/3338906.3340442} of slice-finding emphasizes that even though $\mathcal{M}'s$ MCR $M_2/N_2$, if known, on $D_2$
may be similar to its MCR $M_1/N_2$ in training on $D_1$, that the weak slices found are still important, particularly if some have large support $n_{2,i}$ or very low MCR.  Therefore, detecting changes in the relative slice sizes $n_{j,i}/N_j$ between $j=1,2$ can help detect overall change in the data, without using or knowing $D_2$'s mis-classifications $M_2$.  

\subsection{Hypothesis test setup for dataset comparison}
\label{subsec:hypothesisSetup}

Let $\hat{\pi}_{j,i}=n_{j,i}/N_j$ be the observed size of slice $i$ relative to the size of dataset $j$; its theoretical counterpart is $\pi_{j,i}$.  Both our applications involve conducting $K$ hypothesis tests with null
$H_{0,i}\colon \pi_{2,i}-\pi_{1,i}=0,\: i=1,\dots,K$ based on the observed $\hat{\pi}_{j,i}$, using the normal test for differences in proportions (or its continuity-corrected version in \cite{Y1934}).  Significant changes between datasets in the proportions of records falling into each slice $i$ ($\hat{\pi}_{j,i}$) is likely to indicate significant distributional change in these regions of interest affecting $\mathcal{M}$'s performance.

The p-values $\{p_i\colon i=1,\dots,K\}$ of each test $i$ are then pooled. We would like to use these to make a single decision as to whether the slice mappings indicate that $D_2$ has changed relative to $D_1$, with a statistical guarantee of correctness.  One way is to use the Holm-Bonferroni procedure (\cite{H1979}) which makes an accept/reject (i.e., drift/not drifted) decision on each hypothesis $i$ (slice), adjusting for the fact that we are making multiple tests.  This procedure controls the family-wise error rate (FWER), the probability that \emph{at least one} of the slices will have falsely been detected to have drifted, to be no more than a pre-specified $\alpha$ (e.g., 0.05).  Furthermore, can be used even with dependence between the hypotheses, which is likely to happen if the slices have overlapping region coverage (i.e., if the intersection of two slices grows, they both will as well).  We decide drift has occurred if \emph{any} individual slice hypotheses detects change.  The false positive probability here will be $\alpha$.  We currently weight each slice $i$'s test equally, but plan to experiment with multiple testing adjustments (\cite{BH1997}) that give higher weight to larger slices (based on $n_{1,i}$ or, say, $(n_{1,i}+n_{2,i})/(N_1+N_2)$) or those with higher MCRs ($m_{1,i}/n_{1,i}$). Holm's method may be too sensitive to changes in a very small number of slices, however.

P-values, such as those produced by our hypothesis tests, are well-known to have some drawbacks. For instance, a given `effect' (e.g., the difference between a hypothesized and observed slice proportion, as in our case) can be declared statistically significant if there is a lot of data, even if the effect itself is small.  Also, the p-value has a probabilistic interpretation only assuming the null $H_0$ is true, but the likelihood of this is unknown.  Measures of effect size, such as Cohen's $h$\footnotemark (\cite{cohen_h_book},\cite{PASS_software}), quantify the magnitude of the practical significance of the `effect', without these drawbacks. In future work, we consider using Cohen's $h$ directly, rather than p-values, or perhaps modified in a statistical measure, to declare a slice has significantly changed in size.  However, to our knowledge, effect size measures are not typically adjusted for multiple hypotheses, as in the Holm procedure.

\footnotetext{Cohen's h compares two proportions, in this case $\hat{\pi}_{1,i}$ and $\hat{\pi}_{2,i}$.  The effect size is $h=\arcsin{\sqrt{\hat{\pi}_{2,i}}} - \arcsin{\sqrt{\hat{\pi}_{1,i}}}$.
}


\subsection{Goal 1: Detect data distribution changes that are likely to affect the ML model results} \label{subsec:goal1}

The alternative hypothesis here for each slice $i$ is the two-sided $H_{A,i}\colon\pi_{2,i}-\pi_{1,i}\ne 0$.  Showing that $H_{0,i}$ of equality is rejected consistently across slices indicates there is some change in the data features' distribution in the problematic areas identified by the slices.  Whether or not this results in a change (better or worse) of the observed MCR on $D_2$ relative to $D_1$, the change in the feature distributions in the weak areas, as detected by the slices, may still be important to investigate.







 
\subsection{Goal 2: Detect likely degradation in model performance with no labels, based on data distribution changes} \label{subsec:goal2}

The alternative hypothesis here for each slice $i$ is the one-sided $H_{A,i}\colon\pi_{2,i}-\pi_{1,i} > 0$. 
If the MCR of $\mathcal{M}$ increases from $D_1$ to $D_2$, that is, $M_2/N_2 > M_1/N_2$, then the slices, in which mistakes are concentrated, should grow, rather than shrink, which justifies the use of the one-sided rather than two-sided alternative for this goal.  Thus, observed growth in relative sizes of many slices---that is,  $\hat{\pi}_{2,i}>\hat{\pi}_{1,i}$---reasonably supports the inference that the MCR has increased.

Slices often tend to overlap, however, in their coverage of data records. The overall MCR $M_j/N_j$ (the one of interest) reflects the \emph{total} number of records in the dataset---not each slice---that are mis-classified.  We can imagine cases where, say, the MCR fell but in $D_1$ the overlap of mis-classified observations between slices was high, and in $D_2$ the overlap decreased so that each slice grew in relative size even though the total mis-classifications $M_2$ fell.  However, we believe in general that, particularly with many slices, that this will be unlikely to happen; making inferences from the relative size of slices $\hat{\pi}_{2,i}$ vs $\hat{\pi}_{1,i}$, should be is reasonable, and also much more computationally feasible than calculating overlaps over many slices.





\section{Experiments} \label{sec:experiments}
Section \ref{subsec:datasets} describes the data sets that we used in our experiments.
Section \ref{subsec:randomDataSplits} describes how these data sets were used in the experiments, specifically explaining the random data splits applied. 
Sections \ref{subsec:goal1driftIntroduction} and \ref{subsec:goal2driftIntroduction} define the types of drift injected for each of the detection goals.

\subsection{Data sets and ML models} \label{subsec:datasets}

Our first dataset is \textbf{Adult} (\cite{UCI_adult}). 
This is a loans information dataset which was used to predict whether a person's income exceeds 50K/yr based on information available in the US Census data. The test set has about 14K records and 13 features.
We trained a simple random forest model on this dataset. Weak slices were detected on the original feature set.

Our second dataset is \textbf{Anuran} (\cite{UCI_anuran}), which is used in classification tasks trying to recognize Anuran species (frogs) through their calls. The test set has over 2000 records and 22 continuous features extracted from audio clips. The data also has 3 possible targets: family/genus/species of frog.
We also trained a simple random forest model on this dataset. Weak slices were detected on the original feature set.

Finally, a proprietary dataset \textbf{MP} consisting of about 570 mammography images from a medical provider (MP).
An InceptionResnetV2 with Global Max Pooling model (\cite{Szegedy_Ioffe_Vanhoucke_Alemi_2017}) was trained to classify the images as containing a tumor or not.  A large set of meta-features (e.g., patient race or religion, information about previous tests and biopsies, information about the size and location of the lumps as well as data about the imaging machine), not used in the image classification task, were used for detecting weak slices.

\subsection{Random data splits}
\label{subsec:randomDataSplits}

Each of our three datasets $D$ (see Table~\ref{tab:test_datasets}) contains a Boolean indicator column ($\boldsymbol{\theta}$) of whether each observation was incorrectly classified by the model $\mathcal{M}$ ($D$ served as the test set for $\mathcal{M}$ trained on another dataset).  For each $D$, we create 50 random 50-50 splits, indexed by $b$, into a baseline set $D_1^b$ and deployment set $D_2^b$, of sizes $N_1^b$ and $N_2^b$ (equal up to a difference of 1). The splits are stratified on $\boldsymbol{\theta}$, so $D_1^b$ and $D_2^b$ for each split $b$ have the same MCR.


The slice-finder algorithm was run on each $D_1^b$ independently, giving slice sets $\{S^b\}$; let $K(b)=|\{S^b\}|$ be the number of slices found, which can differ for each $b$.  The mapped slices to $D_2^b$ thus have sizes $n_{2,i}^b,\: i=1,\dots,K(b)$, etc.  These dataset splits and slices were held constant for both of our drift detection goals.

\begin{table}[ht]
\centering
\begin{tabular}{r|rrrr|}
  \hline
Dataset & Features & Records ($N$) & Misclassified ($M$) & MCR \\ \hline
 Adult & 13 & 14,653 & 2,178 & 0.149\\
 Anuran & 22 & 2,159 & 41 & 0.019 \\
 MP & 130 & 568 & 236 & 0.415\\
  \hline
\end{tabular}
\caption{\label{tab:test_datasets}Datasets used in our experiments.}
\end{table}

Table~\ref{tab:slice_statistics} shows summary statistics for each dataset of the slices found on the training set $D_1^b,\:b=1,\dots,50$.  Statistics are the averages across the 50 splits. The slices shown are after a conservative filtering out of slices not satisfying a basic hypergeometric test significance threshold, as in \cite{10.1007/978-3-030-62144-5_6}.

We note first that for all three datasets, typically around 94\% (`\% error coverage') of mis-classified records are included in at least one slice, despite the very different MCRs.  Though 3-way slices were attempted for `Adult' and `Anuran', the hypergeometric threshold filtered all of them out.  In `Anuran', there are many more slices relative to the number of observations and features than in `Adult'.  In `MP', there are relatively few slices, but they tend to be all single-feature slices.  Furthermore, as mentioned in Section~\ref{subsec:datasets}, many of the features have very low variance, and thus may tend to be less helpful individually in identifying mis-classified records; thus, only about 20.5\% of features are even used in any slice.
This is good because it means that the source of mis-classification can be more localized to a small subset of features, and thus $\mathcal{M}$ may be easier to diagnose. However, then our sliced-based drift detection methods are more limited to identifying distribution changes in this smaller feature set.

\begin{table}[ht]
\centering
\begin{tabular}{l|rrr|}
  \hline
  & Adult & Anuran & MP\\ \hline
  Train obs. & 7,326 &	1,079 & 	284 \\
  Num. weak slices & 187 &	771	& 29
\\
  \% 1-feat. slices & 9.57 &	5.63 &	100.00 \\
  \% 2-feat. slices &
  90.43	& 94.37 &	0.00 \\
  \% error coverage & 94.28 &	92.67 &	94.73 \\
  \% features in a slice & 100.00 &	99.42 &	20.47\\
  \% features in 1-feat. slice & 84.15 &	99.23 &	20.47\\
  \% features in 2-feat. slice & 100.00 &	95.45 &	0.00\\ \hline
\end{tabular}
\caption{\label{tab:slice_statistics}Summary statistics of slicing procedure}
\end{table}

\subsection{Goal 1 drift introduced} \label{subsec:goal1driftIntroduction}

Goal 1 (Section~\ref{subsec:goal1}) attempts to use changes, either increase or decrease, in the relative slice sizes $\hat{\pi}_{1,i}$ vs $\hat{\pi}_{2,i}$ to infer general distribution changes in the data region covered by slice $i$. If enough individual slices seem to change in relative size, this may indicate large overall distribution changes (univariate and multi-variate associations) between $D_1$ and $D_2$, particularly in the weak data regions, regardless of the MCRs.

It is common to introduce noise, particularly parametrically-distributed white noise, into observed data to test the robustness of an algorithm.  To avoid parametric assumptions, the procedure first resamples all rows with replacement, which affects univariate feature distributions.  We then introduce drift by swapping row values within a feature column; this can affect multi-way feature associations, including those between numeric and categorical features.  Row swaps can create record combinations, such as $\{\textrm{height}=1.5\textrm{m},\:\textrm{weight}=140\textrm{kg}\}$ or $\{\textrm{years\_education}=12,\:\textrm{age}=10\}$, which are highly improbable or impossible by definition.  However, these are simply extreme examples of distribution change, and a definitional constraint is simply an example of an inter-feature correlation.  Furthermore, since our detection procedure uses only the feature space and not the target label $\mathbf{y}$, the method is not affected if the feature combination values are improbable.



Our procedure is as follows.  Let $D_1^b$ and $D_2^b$ be splits of the same dataset $D$. Assume they have the same feature set $F$ (e.g., $F=\{\textrm{age},\textrm{capital\_gain},\dots\}$), of size $|F|$.  We perform the following permutation distortion on each $D_2^b$ to produce $D_2^{b*}$.
Let $r, c \in (0,1]$ be specified row and column proportions; we then perform distortion permutations on $R=\mathrm{min}(1, \mathrm{round}(rN_2))$ feature rows and $C=\mathrm{min}(1, \mathrm{round}( c|F|))$ feature columns, respectively.  Initially, a unique subset of features $f\subseteq F$, where $|f|=C$, is randomly selected.  Let $I \subseteq \{1,\dots,N_2\}$ be a unique set of row indices, where $|I|=R$.
Our permutation procedure has several parameter settings:

\begin{itemize}
    \item \texttt{keep\_rows\_constant}: if \texttt{True}, a single subset $I$ is selected, and only rows in $I$ are affected in their feature values in $f$.  If \texttt{False}, a new subset $I$ is chosen at random for each feature in $f$.  Within each feature in $f$, the values in the rows indexed by $I$ are randomly permuted. 
    \item \texttt{repermute\_each\_column}: Only applies if \texttt{keep\_rows\_constant=True} (that is, the subset $I$ is fixed across columns in $f$).  If \texttt{False}, a fixed permutation $I'$ of $I$ is made.  Across $f$, the rows indexed in $I$ are exchanged with those in $I'$.  If \texttt{True}, a new permutation $I'$ of $I$ is determined for each feature in $f$.
    \item \texttt{force\_different}: if \texttt{True}, this attempts to ensure that each time a permutation of $I$ to $I'$ is done, that for the given feature, the set of values is changed in at least one row, so that permutation distortion does not result in an output dataset that is identical to the input.
\end{itemize}

\begin{figure}
    \centering
    \includegraphics[scale=0.6]{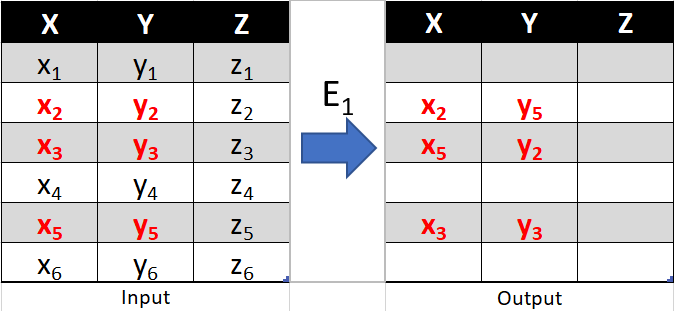}
    \includegraphics[scale=0.6]{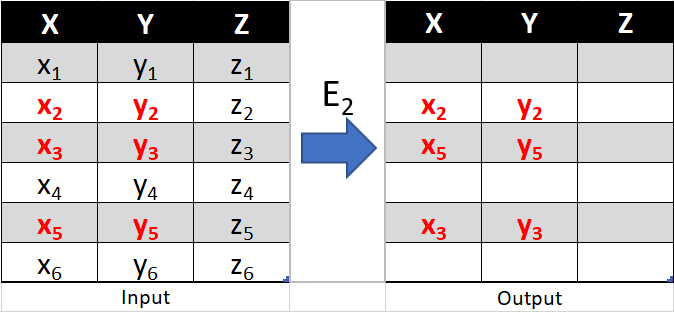}
    \includegraphics[scale=0.6]{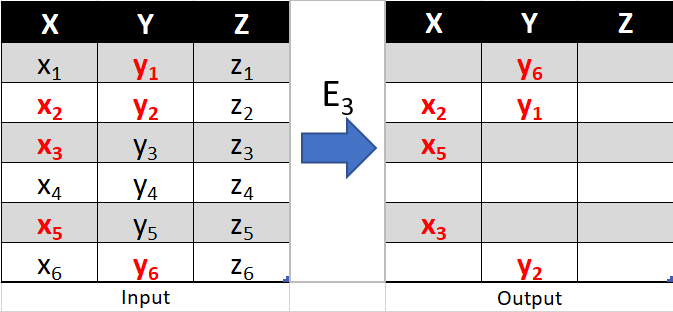}
    \caption{\label{fig:experiment_settings}
    Diagrams illustrating permutation settings $E_1,\:E_2$, and $E_3$ (Table~\ref{tab:permutation experiments} from top to bottom).  A dataset $D_2$ (the left side of each pair) with $N_2=6$ observations and $|F_2|=3$ features is to be permuted and output.  In each, $R=3$ rows and $C=2$ columns ($f=\{\textrm{X},\:\textrm{Y}\}$, but not Z) are selected.  In each, the red (bold) cells are permuted; the output shows only the changed cells.
    In $E_1$ and $E_2$, a single set of row indices $I=\{2,3,5\}$ is randomly selected.  In $E_2$ (middle), a single permutation of $I$, $\{2,5,3\}$ is used for rows $I$ for both columns.  In $E_1$, a new permutation of $I$, $\{5,2,3\}$, is selected for column Y.
    In $E_3$ (bottom), a different set of indices $I'=\{1,2,6\}$ is chosen for the Y column. On each of the columns selected, the row indices $I$ and $I'$ are permuted independently.}
\end{figure}

We thus have three experimental settings, as defined in Table~\ref{tab:permutation experiments} and illustrated in Figure~\ref{fig:experiment_settings}.  Setting both parameters as \texttt{False} is meaningless as it implies no dataset change. 

\begin{table}[ht]
\centering
\begin{tabular}{r|rr|}
  \hline
Experiment & \texttt{keep\_rows\_constant} & \texttt{repermute\_each\_column} \\ \hline
 $E_1$ & True & True \\
 $E_2$ & True & False \\
 $E_3$ & False & True\\ 
  \hline
\end{tabular}
\caption{\label{tab:permutation experiments}Permutation experiment settings}
\end{table}

In settings $E_1$ and $E_2$ (top two of Figure~\ref{fig:experiment_settings}), permutation occurs within fixed block of rows for a subset of columns.  These distort the correlations between these columns and the others for this row subset, by permuting the rows.  The more columns and rows are selected, the higher the potential resulting distortion.  $E_2$ preserves the correlation within the block, while $E_1$ distorts it as well.  $E_3$ causes the most distortion in the feature correlations by changing a different row subset for each features.

For each experimental setting above, five of the 50 splits (see~\ref{subsec:randomDataSplits}) were chosen randomly.  The deployment set $D_2^b$ for each selected split $b$ had its rows resampled with replacement five times independently.  On each of these five, the permutation distortion was performed with each pairwise combination of row and column proportions $r, c \in \{0.1, 0.25, 0.5, 0.75, 1.0\}$, to create a distorted $D_2^{b*}$, to which the slice set $\{S^b\}$ was mapped.  Higher $r$ and $c$ correspond to greater potential distribution drift between $D_1^b$ and $D_2^{b*}$.

For all permutations, we set \texttt{force\_different=True} to ensure the permuted dataset $D_2^{b*}$ is different from $D_2^b$ after its resampling.  The full MP dataset has 130 columns, many of which are binary-valued and imbalanced (one of the values has frequency $>400$ out of 568), and are thus problematic.  This poses a computational bottleneck because if the block of data to be permuted include many of these problematic columns, then it may be impossible, or require many attempts, to ensure that the permuted result is actually different, if we set \texttt{force\_different=True}.
Therefore, for the this experiment on MP only, we first drop many of the problematic columns, leaving 69, before building the slices in the first place.  We note that we need to do this to ensure our procedure actually creates a distortion in $D_2^{b*}$; it is not a weakness of our drift detection method, because in reality if a theoretical permutation didn't actually change the data at all, this is the same as no permutation at all.

\subsection{Goal 2 drift introduced} \label{subsec:goal2driftIntroduction}

Goal 1 detects general feature association changes, simulated here by permutations, through change in the relative sizes of slice sets; general drift may or may not affect the MCR.
Goal 2 specifically tries to infer whether the MCR has changed between $D_1$ and $D_2$, particularly if it increases.


Our experimental procedure creates a distorted test dataset $D_2^*$ of the same size $N_2$ as $D_2$, from $D_2$ but with known $M_2^*$ misclassifications; if $M_2^*>M_2$, the MCR increases.  All records in $D_2$ are resampled with equal weight, but with different total draws of incorrect or correct classifications (which are known), to obtain the desired totals $M_2^*$ and $N_2-M_2^*$ in $D_2^*$. If $\phi=\frac{M_2}{N_2-M_2}=\frac{\textrm{MCR}}{1-\textrm{MCR}}$ is the odds ratio of incorrect to correct classifications in $D_2$, $M_2^*$ is set\footnotemark~ so that the corresponding odds ratio $\phi^*=\frac{M_2^*}{N_2-M_2^*}$ on $D_2^*$ satisfies $\phi^*=k\phi$ for $k>0$, where $k>1$ increases the MCR.
We use the approach of scaling $\phi$ rather than the MCR because the MCR is bounded in $[0,1]$, and it is easier to formulate multiplicative increases in $\phi$, which increase the MCR non-linearly, than in the MCR itself.


\footnotetext{Set  $M_2^*=\mathrm{max}\left(\mathrm{min}\left(\mathrm{round}\left(\frac{k M_2 N_2}{N_2-(1-k)M_2}\right), N_2\right), 0\right)$.}



Unlike the permutations approach of Goal 1, the resampling of records \emph{does} change the univariate distribution of the features, so a goodness-of-fit test, without the use of slices, can detect the change from $D_1$ to the resampled $D_2^*$, but this cannot easily tell us if the MCR has changed or how.  However, because the slices specifically target the weak data areas where the MCR is higher, we \emph{can} infer that the MCR has changed, particularly that it has increased.  Of course, our technique ignores changes in features that are not part of weak data slices, which are less important if they don't much affect $\mathcal{M}$'s performance.  However, changes in these features can be detected indirectly by changes in slices for other features that are correlated with them. 

Here, we have increased the MCR by duplicating records from $D_2^b$ in each $D_2^{b*}$. It may be more realistic to, say, use  generative adversarial networks (GANs,~\cite{GPMXWOB2014}) to generate a synthetic $D_2^*$ very similar to $D_1$, with higher MCR but without duplicating records.  This is a topic for future research.



\section{Results} \label{sec:results}
For each of the goals described in Section \ref{sec:methodology} we show the experimental results that demonstrate the feasibility of our method in achieving the goal. The data sets we used and the experimental setting are described in Section \ref{sec:experiments}.

\subsection{Goal 1: Detecting data distribution changes that are likely to affect the ML model results} \label{subsec:goal1results}

We note that the procedure described in Section~\ref{subsec:goal1driftIntroduction} causes distributional change by permuting values within columns in random row subsets.  Since the rows and columns are random, the drift does not specifically target the weak areas covered by the slices.  
Any scenario where distortion was induced ($C,R\geq 1$, except for $c=1$ for the $E_2$ setting, as mentioned below) is considered distribution drift from $D_1^b$ to $D_2^{b*}$, even if the areas covered by slices weren't affected.

\begin{figure}
    \centering
    \includegraphics[scale=0.4]{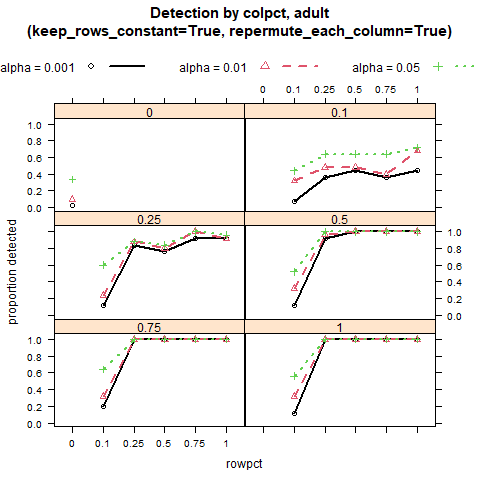}
    \includegraphics[scale=0.4]{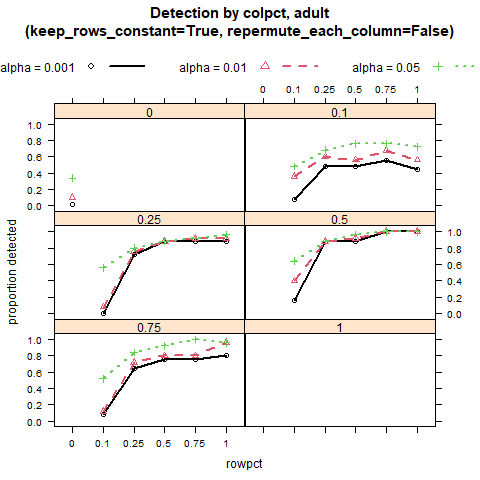}
    \includegraphics[scale=0.4]{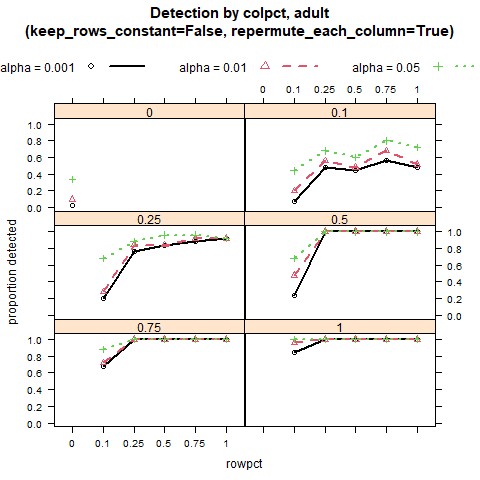}
    \caption{\label{fig:goal1figs_adult} Proportion of times drift was detected on the test set split, over five randomly selected splits $b$ and repetitions of a given permutation setting $c,r$.  Each panel shows a given value of column proportion $c$ (`colpct'), within which the row proportion $r$ (`rowpct') increases.  Plots from top to bottom represent settings $E_1$, $E_2$, and $E_3$ (Table~\ref{tab:permutation experiments}) on the `Adult' dataset.}
\end{figure}

Figure~\ref{fig:goal1figs_adult} shows results of permutation experiments $E_1$, $E_2$, and $E_3$ (top-to-bottom) on the deployment dataset $D_2^b$ of five randomly-chosen splits $b$ of the Adult dataset.  Each panel shows 
the proportion of times the drift was detected, for each $r,c$ setting, for a total of 25 comparisons (five initial splits, and five re-samplings of each of them).  The panels show, for a fixed $c$, the proportion of times the drift (permutation of rows/columns) was detected, for increasing $r$, at three different levels $\alpha$ (line plots).  Higher combination of $r$ and $c$, which correspond to more rows and columns being permuted, cause more distortion, which are more likely to be detected, at any given experimental setting.

Note that in the first panel of each, when $c=r=0$, five instances of $D_2^{b*}$ are the rows of $D_2^b$ resampled with replacement with no permutation. A true `no drift' comparison is $D_1^b$ vs its original test set $D_2^b$, since they are essentially from the same data distribution. The resampling will tend to create some duplicate records from $D_2^b$ in $D_2^{b*}$, while dropping some records, so it is likely to create some slight distribution change. This is why the detection rates in the first panel are higher than the respective $\alpha$ in each case (they would tend to be at $\alpha$ if this represented a true false positive).

However, the first two experimental settings $E_1$ and $E_2$, for which the initial rows selected ($I$) are held constant across the selected rows (\texttt{keep\_rows\_constant=True}), means that the permutations across the columns affect a more limited row subset and preserves more of the between-feature correlation within these selected rows $I$, particularly if \texttt{repermute\_each\_column=False} as well, as in $E_2$.  This more limited distortion is less likely to be detected than $E_3$, where the selected rows to be permuted change for each column.  If $C=1$ (for `Adult', since $|F|=13$ features, this occurs for $c=0$ or $0.1$ in our experiments), only one column is permuted, and all experimental settings are equivalent.  However, in the bottom plot for $E_3$, it is clear that the higher distortion is detected with greater probability than the others at any given $r,c$, for $c\geq 0.25$.  For $E_2$ (middle), the panel for $c=1$ is omitted because this represents a simple permutation of the rows in the dataset.

\begin{figure}
    \centering
    \includegraphics[scale=0.4]{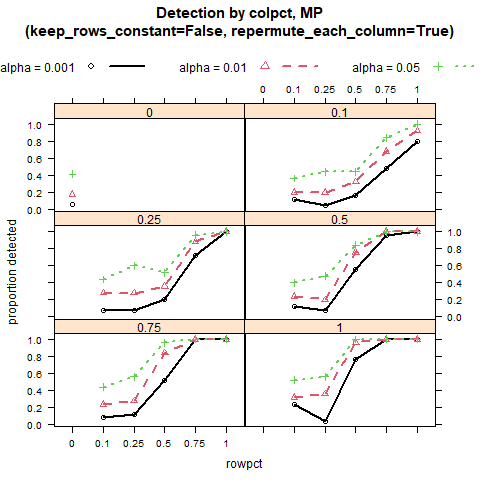}
    \caption{\label{fig:goal1figs_mp} Detection success results for $E_3$ on the reduced MP dataset.}
\end{figure}

As mentioned in Subsection~\ref{subsec:goal1driftIntroduction}, for the MP dataset, we reduced the number of features to 69 before building slices and conducting permutations.  For MP, the differences between experiments were similar but smaller in magnitude than for the Adult dataset.  For $E_3$, while drift was detected more often than in $E_1$ or $E_2$ at various settings, for instance, the detection proportion for $E_3$ was both lower absolutely and was less sensitive to changes in $r,\:c$ than for Adult.  Results on the Anuran dataset were similar to MP.  This is shown in Figure~\ref{fig:goal1figs_mp}.

\subsection{Goal 2: Detecting degradation in model performance with no labels} \label{subsec:goal2results}

For each test dataset $D_2^b$, $b=1,\dots,50$, distorted datasets $D_2^{b*}$ are created by re-balancing the ratio of correct to incorrectly-classified records  (Section~\ref{subsec:goal2driftIntroduction}) for multipliers $k=1.0,\: 1.25,\: 1.5,\: 1.75,\: 2.0,\: 3.0,\: 5.0,\: 7.5,\: 10.0$ of the calculated odds ratio $\phi$ of $D_2^b$.  For each $b$, each slice in $\{S^b\}$ from $D_1^b$ was mapped to $D_2^{b*}$. A one-sided continuity-corrected test was conducted for each slice $i$ to decide between $H_{0,i}\colon\: \pi_{2,i}-\pi_{1,i}=0$ vs the alternative $H_{A,i}\colon\:\pi_{2,i}-\pi_{1,i}>0$, where $H_{A,i}$; the one-sided alternative implies the MCR is likely to be increasing, specifically (see Section~\ref{subsec:goal2driftIntroduction}).

\begin{figure}
    \centering
    \includegraphics[scale=0.3]{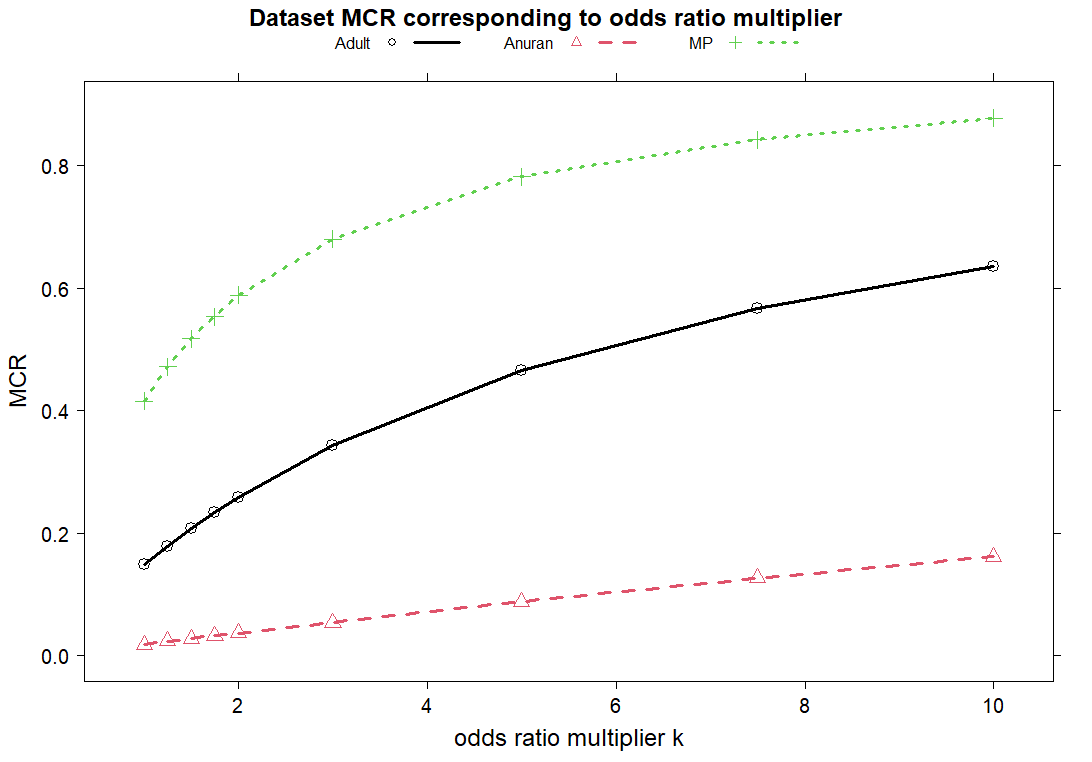}
    \caption{\label{fig:multiplier2mcr}Multiplier $k$ and corresponding MCR for each dataset.}
\end{figure}

\begin{figure}
    \centering
    \includegraphics[scale=0.45]{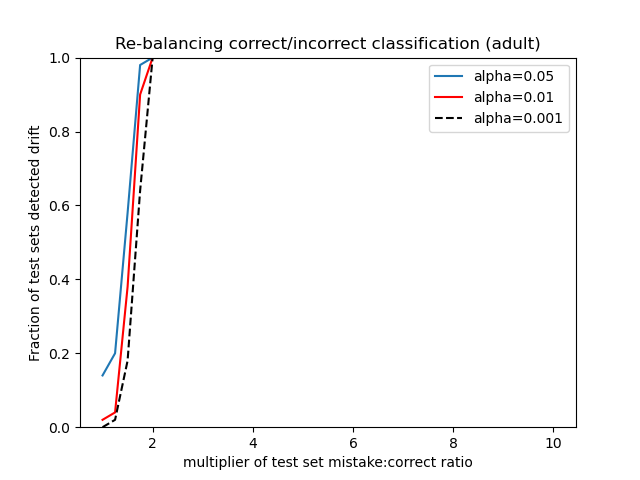}
    \includegraphics[scale=0.45]{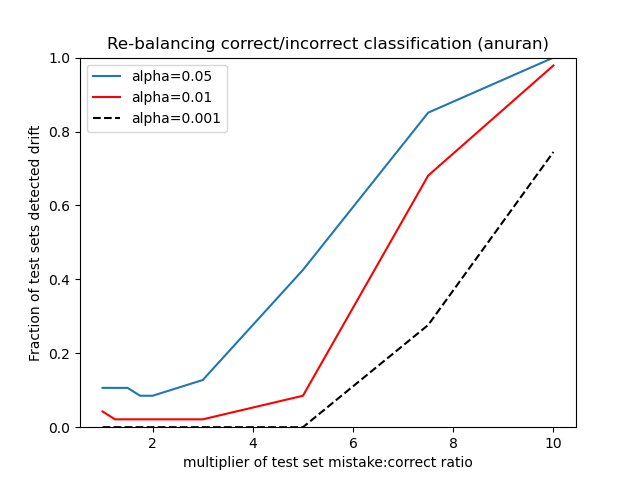}
    \includegraphics[scale=0.45]{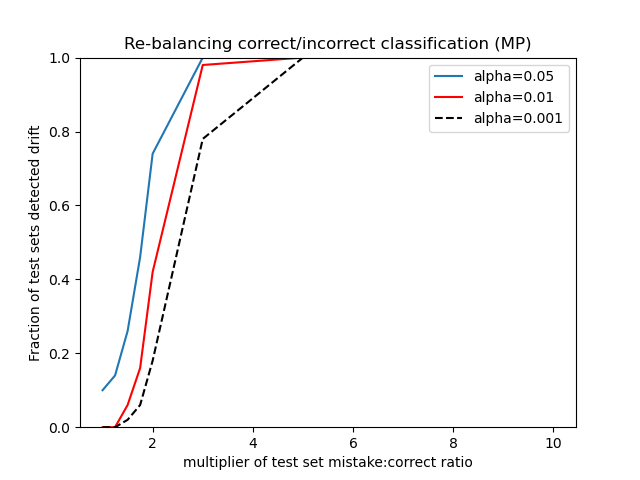}
    \caption{\label{fig:goal2figs}Fraction of test set splits detected as having higher MCR (Section~\ref{subsec:goal2results}), at various $\alpha$ thresholds and ratio multipliers $k$.}
\end{figure}

For a given dataset (Adult, etc.), we perform one re-balancing of each split $b$ at each multiplier $k$. The full MP dataset was used here, rather than the reduced on as in Section~\ref{subsec:goal1driftIntroduction}.  Figure~\ref{fig:multiplier2mcr} shows the MCR corresponding to a given multiplier $k$, assuming a perfect 50-50 split where $M_2=M/2$ and $N_2=N/2$.  Since the datasets have different initial MCRs (for $k=1.0$), the absolute level of MCR at each $k$ for the different datasets differ, and it's likely that larger absolute changes in MCR are easier to detect. Figure~\ref{fig:goal2figs} thus shows the proportion of the 50 splits for which the re-balancing was detected, at each multiplier $k$ above. The fact that Anuran (middle) has the lowest initial MCR is likely responsible for the fact that the multiplier $k$ must be much higher than the other datasets for the change in MCR to be detected. 

\section{Related work} \label{sec:related}

Traditional statistical methods have long utilized permutation-based tests. In particular, restricted randomization techniques (\cite{good1994permutation})  inspired many ML  assessments, such as permutation-based assessment of classifiers performance (\cite{permutationTests}) or concept drift detection via random permutations of the examples (\cite{Harel2014ConceptDD}). We utilize permutation-based restricted randomization techniques to methodologically define data drift. Concept drift has long been an area of active research (\cite{Tsymbal2004ThePO}). However, much of the concept drift work is focused on developing a way to automatically update an ML model in the presence of drift or focuses on tracking performance indicators for drift detection (\cite{Klinkenberg98adaptiveinformation, wk-lpcdhc-96}). \cite{outlierdetect} proposes sequential analysis for drift detection. 

Our drift detection technique is based on data slicing. such as that implemented by FreaAI  (\cite{10.1007/978-3-030-62144-5_6, 10.1145/3338906.3340442}). We use the FreaAI data slices to define a trigger mechanism for alerting on drift. Other approaches use different triggers; for example \cite{lanquillon99} utilizes information filtering, and \cite{classifierConf} utilizes classifier confidence distribution changes. 

Drift detection is theoretically possible by capturing the data distribution and alerting on significant changes. Unfortunately, this is very challenging for data with many dimensions that is drawn from an unknown (non-parametric) distribution  (see \cite{Scholkopf:2001:ESH:1119748.1119749}). In general, as the dimensionality increases so does the likelihood of having insufficient data for reliable statistical estimation, but Maximum Mean Discrepancy (\cite{Gretton:2012:KTT:2188385.2188410}) or Wasserstein distance (\cite{OLKIN1982257}) are examples of progress in this area.

\section{Conclusion} \label{sec:discussion}
We propose a novel approach for drift detection that utilizes weak data slices and defines a distribution based on them.  The data drift types that our method is effective in detecting are either (1) distributional change or (2) increases in mis-classifications, occurring in the data regions identified by the weak slices.
We demonstrate that our method effectively finds such drift even when the underlying  univariate feature distribution is unchanged. Our method also naturally ignores any changes in features that are not in any weak data slice.  Future work will investigate the relation between drift magnitude and risk resulting from ML model degradation, as well as use of effect size rather the p-values in our test (see Section~\ref{subsec:hypothesisSetup}).


\section{Data Availability} \label{sec:data_availability}

We are not making the code and data available at this time since one of them is proprietary.  Also, the code is part of a product under development.
\bibliographystyle{abbrv}
\bibliography{main} 

\end{document}